\pdfoutput=1
\documentclass[10pt,twocolumn,letterpaper]{article}
\usepackage[pagenumbers]{cvpr}
\usepackage{booktabs}
\usepackage[T1]{fontenc}
\usepackage{bm,array,multirow,amsthm}

\newcommand{\dv}{\bm{d}}           %
\newcommand{\ov}{\bm{o}}           %

\theoremstyle{plain}
\newtheorem{proposition}{Proposition}

\makeatletter
\newbox\kc@pendinglabels
\@ifundefined{@ShipoutPicture}{}{%
  \let\kcorig@ShipoutPicture\@ShipoutPicture
  \renewcommand{\@ShipoutPicture}{%
    \global\let\kc@if@inlabel\if@inlabel
    \global\setbox\kc@pendinglabels\box\@labels
    \global\@inlabelfalse
    \kcorig@ShipoutPicture
    \global\setbox\@labels\box\kc@pendinglabels
    \global\let\if@inlabel\kc@if@inlabel}}
\makeatother

\definecolor{cvprblue}{rgb}{0.21,0.49,0.74}
\usepackage[pagebackref,breaklinks,colorlinks,allcolors=cvprblue]{hyperref}
\definecolor{mnA}{HTML}{2B8A3E}\definecolor{mnInk}{HTML}{1B1F2A}
\hypersetup{pdftitle={FirstCross: Convex Single-Photon 3D Reconstruction from Sparse Views},pdfauthor={Chi-Wei Lee, Cheng-Han Chiang, Yi-Chun Chang, An-Kai Chen, Po-Chih Kuo}}
\def\confName{CVPR}
\def\confYear{2027}

\newcommand{\xb}{\bar{x}}
\newcommand{\ldn}{\ell^{\downarrow}}
\newcommand{\lup}{\ell^{\uparrow}}
\newcommand{\tst}{t^{\star}}
\newcommand{\that}{\hat{t}}
\newcommand{\pos}[1]{[#1]_{+}}

\title{FirstCross: Convex Single-Photon 3D Reconstruction from Sparse Views}
\author{Chi-Wei Lee \quad Cheng-Han Chiang \quad Yi-Chun Chang \quad An-Kai Chen \quad Po-Chih Kuo\\
National Tsing Hua University, Hsinchu, Taiwan\\[2pt]
{\small\href{https://arthur031221.github.io/firstcross/}{\textcolor{mnA}{\texttt{arthur031221.github.io/firstcross}}}}}

\begin{document}
\twocolumn[{%
\renewcommand\twocolumn[1][]{#1}%
\maketitle
\begin{center}
\vspace{-1.2em}
{\setlength{\unitlength}{\dimexpr\textwidth*100/276000\relax}%
\begin{picture}(2760,1760)
\put(0,0){\includegraphics[width=\textwidth]{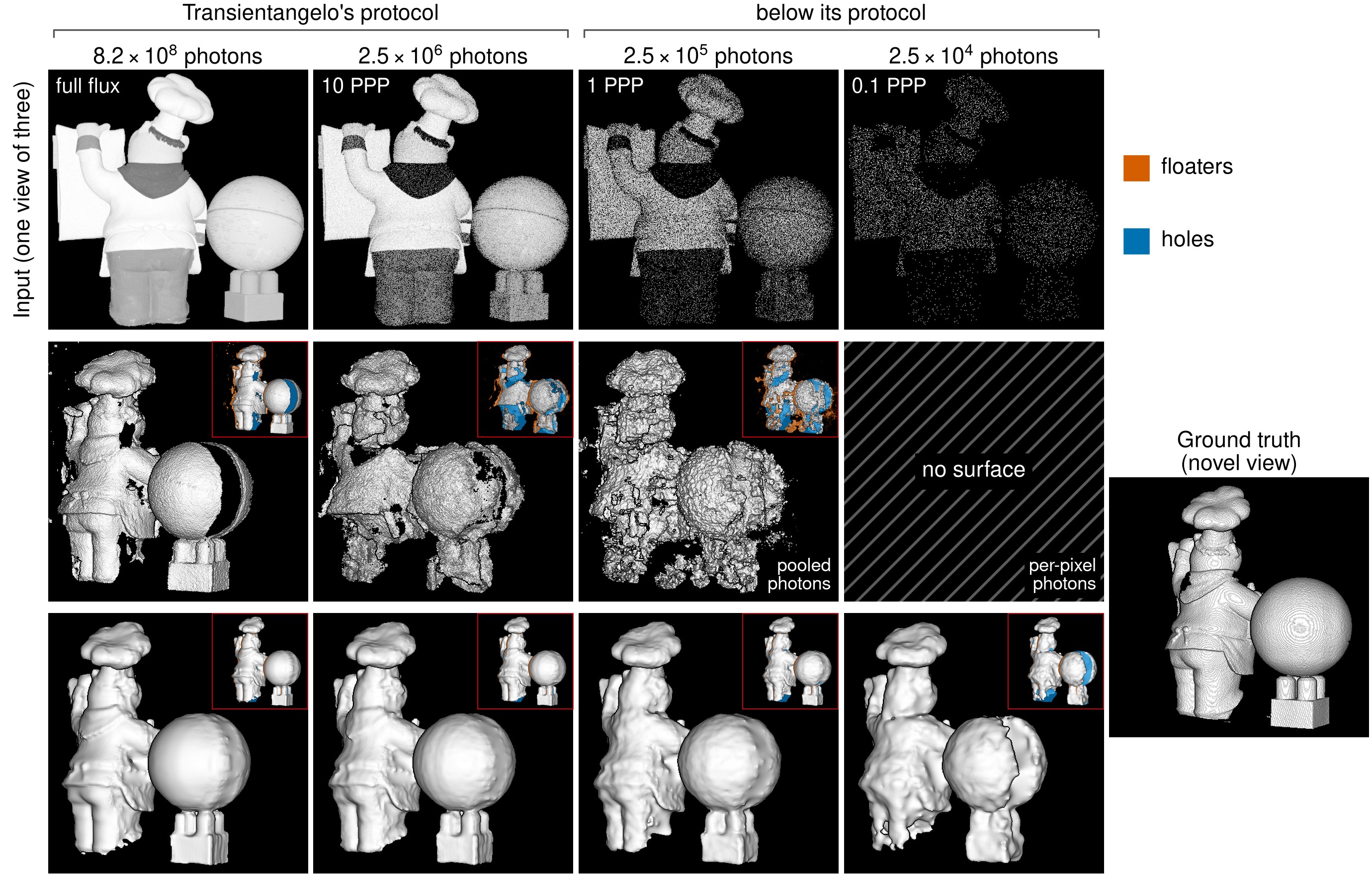}}%
\put(57.5,812){\rotatebox{90}{\makebox[0pt][c]{\fontencoding{T1}\fontfamily{phv}\fontseries{m}\fontshape{n}\fontsize{8.0}{9.6}\selectfont Transientangelo~\cite{luo2025transientangelo}}}}%
\put(57.5,265.5){\rotatebox{90}{\makebox[0pt][c]{\fontencoding{T1}\fontfamily{phv}\fontseries{m}\fontshape{n}\fontsize{8.0}{9.6}\selectfont FirstCross (Ours)}}}%
\end{picture}}
\captionof{figure}{\looseness=-1 \textbf{FirstCross reconstructs 3D shape from very few photons.} Novel view of the real test scene chef from three views, with their total photon count above each column. PPP is the mean photon count of an occupied pixel, and 1 and 0.1 PPP are thinned from 10 PPP. Red-framed insets mark depth errors over 2\,cm, as floaters (in front of the true surface or off the object) or holes (behind it or missing). Transientangelo leaves floaters and holes even at full flux. As it recovers no surface from per-pixel photons below its protocol's lowest level, 10 PPP, we show it at 1 PPP with adapted settings (Sec.~S3) on photons pooled over $4\times4$ pixels.}
\label{fig:teaser}
\end{center}
\vspace{0.5em}
}]

{\let\thefootnote\relax\footnotetext{\raggedright Emails: {\tt \{levi74108520963, faidavid7, changg587, kylechen.ankai\}@gmail.com, kuopc@cs.nthu.edu.tw}}}
\begin{abstract}
\looseness=-1 %

Single-photon lidar records the arrival time of each returning photon, enabling 3D reconstruction from few views and scarce photons. However, leading neural methods such as Transientangelo require over an hour of non-convex per-scene optimization, and their accuracy degrades as photon flux decreases.
We propose FirstCross, a network-free approach that instead solves a short sequence of convex problems on a voxel grid. For a solid object, the Poisson likelihood of a ray with given signal and background rates depends on the scene solely through its first crossing, the point where the ray first enters the object.
Decomposing this likelihood yields two tractable terms. The first penalizes surfaces in front of detected photons and admits a tight convex closure, namely volume rendering with a running-maximum transmittance in place of NeRF's product transmittance. The second accounts for photons that no surface explains and admits a convex upper bound once the object is assumed to be thick behind each detected surface point.
\mbox{Compared} with \mbox{Transientangelo} under its own \mbox{real-data} protocol, FirstCross achieves lower depth error across every test scene, view count, and flux level. With three views and 100 times fewer photons than at the protocol's lowest flux, our mean object depth error stays under half of Transientangelo's with all photons. Given our detected surface points, the best classical fusion has 1.3 to 1.8 times our three-view error across flux levels. On real room-scale captures with 1\% of their photons, TransientNeRF has 2.8 to 7.7 times our error. FirstCross lets few-view single-photon scanners cut exposure to 1\% and reconstruct an object in about a minute on one GPU.

\end{abstract}

\section{Introduction}
\label{sec:intro}

\looseness=-1 In single-photon lidar, a pulsed laser and a single-photon avalanche diode (SPAD) record a transient histogram of photon arrival times at each pixel. Each signal photon measures range, and a few of them locate a surface to within the width of the instrument response function (IRF), the timing jitter of laser and detector. However, fast acquisition, long range or low-power sensors yield fewer than one photon per pixel~\cite{kirmani2014first,shin2015photon}, as at 0.1 photons per occupied pixel (PPP) in \cref{fig:teaser}.

\looseness=-1 Prior work follows two main routes. The classical route detects a return on each ray (the arrival time of the echo from the surface) and fuses the returns~\cite{kazhdan2013screened} and the free space in front of them~\cite{curless1996volumetric,zach2007tvl1} across views. A ray without a return contributes nothing, although the absence of photons is evidence of free space. The strongest neural methods, transient neural radiance fields~\cite{malik2023transientnerf,malik2024flying} and their surface variant Transientangelo~\cite{luo2025transientangelo}, instead volume-render histograms from a neural field and fit the measurements bin by bin. They are accurate with many photons, but fitting each scene takes over an hour and the objective is non-convex.

\begin{figure*}[t]
\centering
\includegraphics[width=\textwidth]{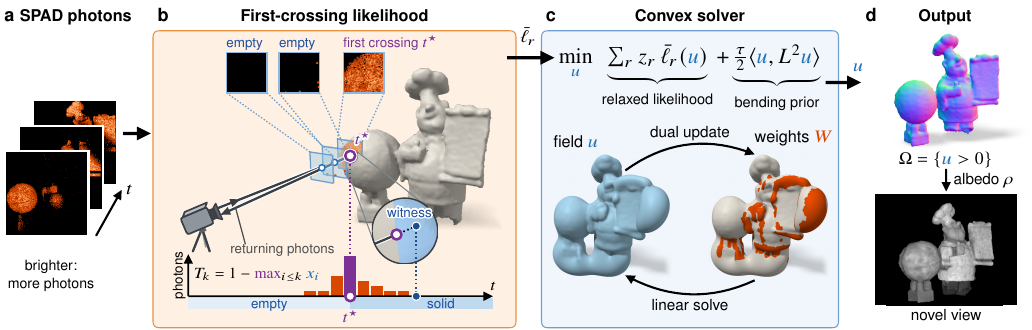}
\caption{\looseness=-1 \textbf{Overview of FirstCross on real data.} FirstCross fits a field~$u$ by a short sequence of convex problems. (a)~Photons of one view at 1~PPP in three arrival-time slices. (b)~The likelihood of ray~$r$ depends on the scene only through its first crossing~$\tst$ (purple). Cross-sections across~$r$ (top) show almost no photons in free space, and the pooled photons of~$r$ (bottom) peak at~$\tst$. The witness (dark blue), just behind the return, is assumed to lie inside the object. (c)~Each of the 12 iterations puts the weights~$W$ (orange) where rays enter the current object, then updates~$u$ by a linear solve. The weight~$z_r$ of ray~$r$ falls with its photon count. (d)~The shape $\Omega=\{u>0\}$ and a novel view.}
\label{fig:pipeline}
\end{figure*}

\looseness=-1 We propose FirstCross, which starts instead from the photon likelihood (\cref{fig:pipeline}). Our key observation is that, for a solid object, photon arrivals along a ray with given signal and background rates depend on the scene only through one point: the first intersection of the ray with the object, which we call the \emph{first crossing}. We write the Poisson likelihood of every ray, with or without photons, as a function of its first crossing and then of the occupancies of its samples in an implicit field on a voxel grid.
\looseness=-1 This likelihood is combinatorial in the occupancies, but it splits into two parts. The \emph{too-early} part, where a surface blocks the path of a photon, is submodular in the set of occupied samples. Its tight convex closure, the largest convex function that matches it on binary occupancies, has a closed form: volume rendering whose transmittance is one minus the running maximum of occupancy along the ray, in place of NeRF's product transmittance. The \emph{too-late} part, where no surface explains a photon, is supermodular and the only source of non-convexity. It requires the ray to enter the object before the photon arrives without specifying where. Our \emph{thickness assumption}, that objects are thicker than a few IRF widths, supplies the location, because the point just behind a detected return then lies inside the object. Each solver iteration therefore works on a convex problem, and the minimum of the first problem does not depend on the initial grid, although our solver reaches it only approximately. With the product transmittance, by contrast, an optimisation started from an occupied grid can leave free space filled. We make three contributions:
\begin{itemize}\setlength{\itemsep}{1pt}
\item\looseness=-1 We show that the first-crossing Poisson likelihood splits into a submodular too-early part and a supermodular too-late part. We derive the tight convex closure of the too-early part in closed form, which is volume rendering with a running-maximum transmittance.
\item\looseness=-1 We derive a convex upper bound on the too-late part from the thickness assumption. Our solver needs no neural network and reconstructs an object in about a minute.
\item\looseness=-1 We demonstrate lower depth error than Transientangelo on every test scene, view count and flux level of its own real-capture protocol. From three views, our error on the object stays below Transientangelo's at full flux even with inputs thinned 100-fold below its lowest flux of 10~PPP.
\end{itemize}
\noindent\looseness=-1 \textbf{Scope.} We assume solid objects, known camera poses and negligible pile-up, which biases detections toward early time bins at high flux. Thin sheets, partially transmissive layers~\cite{wen2026mixed} and mixed pixels at silhouettes are not modelled.

\section{Related Work}
\label{sec:related}

\par\noindent\textbf{Neural transient fields.}
\looseness=-1 TransientNeRF~\cite{malik2023transientnerf} introduced volume rendering of multi-view transient histograms. Transientangelo~\cite{luo2025transientangelo} then targeted sparse views and low flux with a neural signed distance field~\cite{li2023neuralangelo} and regularisers. Other work models propagating light~\cite{malik2024flying,malik2025inverse}, outdoor scenes~\cite{scheuble2025lasso}, low-cost sensors~\cite{mu2024lowcost} and mixed returns~\cite{wen2026mixed}, or learns to reconstruct from a single transient~\cite{li2025trt}, and SP-TransientBench~\cite{sptbench2026} provides room-scale captures.

\par\noindent\textbf{Photon-efficient depth imaging.}
\looseness=-1 First-photon imaging~\cite{kirmani2014first} and later work~\cite{shin2015photon,rapp2017few,heide2018sub,lindell2018single,peng2020photon,tachella2019real} use Poisson statistics for single-view depth, with spatial priors over neighbouring pixels. With several views, all first crossings lie on one surface, so our field couples rays across views.

\par\noindent\textbf{Range fusion and implicit surfaces.}
\looseness=-1 Volumetric fusion~\cite{curless1996volumetric} and its robust convex forms~\cite{zach2007tvl1} combine per-ray distances. Poisson-type reconstruction~\cite{kazhdan2006poisson,kazhdan2013screened,calakli2011ssd,peng2021sap} and radial basis functions~\cite{carr2001rbf} fit an implicit function to oriented points, and variants estimate orientation~\cite{lin2022pgr,hou2022ipsr} or add visibility envelopes~\cite{kazhdan2020envelope}. Our bending prior follows~\cite{calakli2011ssd,carr2001rbf}, but our data term lets zero-photon rays carve free space and weights rays by effective photon counts.

\par\noindent\textbf{Visibility in surface reconstruction.}
\looseness=-1 Space carving~\cite{kutulakos2000carving} and continuous multi-view stereo~\cite{kolev2009continuous} use per-voxel photo-consistency, and $s$--$t$ cuts~\cite{labatut2009robust} and VASE~\cite{tagliasacchi2011vase} use soft free-space constraints from points. Closest to ours are ray potentials~\cite{liu2010ray,savinov2015ray,ulusoy2015ray}, whose cost depends on a ray's first occupied voxel and is minimised discretely or by a convex relaxation with a non-convex visibility constraint~\cite{savinov2016semantic}. Concurrently, Lenain \etal~\cite{lenain2026visibility} regress the first intersection of a 1-Lipschitz network, a non-convex problem. We share this first-crossing structure and submodular tools~\cite{lovasz1983submodular,bach2013learning,narasimhan2005submodular}, but our cost is the photon likelihood, and we derive the tight convex closure of its too-early part and a convex bound on its too-late part. This closure is a running-maximum rendering that models visibility in a convex functional.

\par\noindent\textbf{Volume rendering of surfaces.}
\looseness=-1 NeRF~\cite{mildenhall2020nerf} composites independent opacities. NeuS~\cite{wang2021neus}, VolSDF~\cite{yariv2021volsdf} and UNISURF~\cite{oechsle2021unisurf} derive them from an implicit surface but keep the product along the ray. NeuS matches the running maximum on rays that cross one surface. IDR~\cite{yariv2020idr} renders the first intersection alone, and non-exponential transmittance~\cite{vicini2021nonexp,speierer2026nonexp} and stochastic geometry~\cite{miller2024objects} model opaque surfaces. All of these agree on binary occupancy and differ on fractional values. The running maximum is convex, and \cref{thm:closure} shows that no convex extension is tighter.

\section{From Photons to a Convex Functional}
\label{sec:method}

\looseness=-1 From a few transient histograms with known poses, we seek a field $u$ on a voxel grid that is positive exactly inside the object. We write the likelihood of a ray through its first crossing, replace its too-early and too-late parts with convex surrogates, and minimise the resulting functional (\cref{fig:pipeline}). Sec.~S1 gives proofs and derivations.

\subsection{First-Crossing Likelihood}
\label{sec:ray}

\looseness=-1 Let $\bm{r}(t)=\ov+t\dv$ be a camera ray, with time measured as one-way range, and let its first crossing $\tst$ be the first depth where $u>0$, or infinity if there is none. Neglecting pile-up, photon arrivals on the ray form a Poisson process with intensity $a\,h(t-\tst)+b$, where the IRF $h$ is a Gaussian of standard deviation $\sigma$, $a$ is the expected signal count and $b$ is the background rate. Keeping only the larger of $a\,h$ and $b$ in the log-intensity of each photon changes its cost by at most $\log 2$. Up to constants, the negative log-likelihood of the arrival times $t_1,\dots,t_n$ then becomes
\begin{equation}
\ell(\tst)=a\,\mathbf{1}[\tst<\infty]+\sum_{j=1}^{n}\min\Bigl\{\frac{(t_j-\tst)^2}{2\sigma^2},\ \omega\Bigr\},
\label{eq:ell}
\end{equation}
\looseness=-1 where $\omega=\log(a\,h(0)/b)$, the cost in nats of an unexplained photon, is held constant. A crossing before the arrival of a photon blocks its path, a later or missing crossing leaves the photon unexplained, and a crossing near the photon incurs a quadratic cost. A ray with no photons pays its expected signal count for any crossing and thus carves free space.

\subsection{Convex Closure of the Too-Early Cost}
\label{sec:split}

\looseness=-1 To optimise this cost, we sample the ray at depths $\xi_1<\dots<\xi_K$ and let the binary vector $x$ mark the samples inside the object. The cost depends on $x$ only through the first occupied sample. It is thus a set function $F(A)=\ell(\min A)$ of the set $A$ of occupied samples, where $\ell(k)$ is the cost of a crossing at sample $k$, and $K+1$ means no crossing. If $\ell$ is non-increasing on $\{1,\dots,K+1\}$, then $F$ is submodular, $F(A)+F(B)\ge F(A\cup B)+F(A\cap B)$, and if $\ell$ is non-decreasing, then $F$ is supermodular.

\looseness=-1 Every such cost is the sum of a non-increasing and a non-decreasing part, and we split \cref{eq:ell} one photon at a time,
\begin{equation}
\begin{aligned}
\ldn(t)&=a\,\mathbf{1}[t<\infty]+\textstyle\sum_j\mathbf{1}[t<t_j]\,\phi_j(t),\\
\lup(t)&=\textstyle\sum_j\mathbf{1}[t>t_j]\,\phi_j(t).
\end{aligned}
\label{eq:split}
\end{equation}
\looseness=-1 Here $\phi_j$ is the cost of photon $j$ in \cref{eq:ell}. The too-early cost $\ldn$ penalises a surface on the ray before $t_j$, whereas the too-late cost $\lup$ penalises a missing surface by $t_j$. By monotonicity, $\ldn$ makes $F$ submodular and $\lup$ supermodular.

\begin{proposition}[Tight convex closure]
\label{thm:closure}
Let $\ldn$ be non-increasing with $\ldn(K+1)=0$, $F(A)=\ldn(\min A)$, $\Delta_k=\ldn(k)-\ldn(k+1)\ge 0$ and $\xb_k=\max_{i\le k}x_i$, the running maximum of $x$. Then the Lov\'asz extension of $F$ is
\begin{equation}
F^{L}(x)=\sum_{k=1}^{K}\Delta_k\,\xb_k,\qquad x\in[0,1]^K.
\label{eq:closure}
\end{equation}
It is convex, equals $F$ on binary $x$, and is the largest convex function on $[0,1]^K$ with these values.
\end{proposition}

\looseness=-1 The closure follows from the Lov\'asz extension formula~\cite{lovasz1983submodular,bach2013learning}, which averages the set function over the level sets of the occupancies. A level set starts at or before a sample exactly when the running maximum at that sample reaches the level, and the closure is the expected too-early cost of the first crossing of a random level set. It stays convex when the occupancies are a convex, non-decreasing function $g$ of the field. As clipping $g$ at one would break convexity, occupancies may exceed one inside the object. This costs nothing on rays with a return, where we drop the expected signal count, and the too-early cost stops decreasing behind the last photon (Sec.~S1). The closure is tight for one ray and, summed over rays, a convex surrogate in $u$.

\par\noindent\looseness=-1 \textbf{Running-maximum transmittance.} For $x\in[0,1]^K$, let $T_k=1-\xb_k$, $T_0=1$ and $\pi_k=T_{k-1}-T_k$. Then $F^L(x)=\sum_k \pi_k\,\ldn(k)$, which is volume rendering with transmittance one minus the running maximum. NeRF's transmittance $T_k=\prod_{i\le k}(1-x_i)$ and ours are two extensions of one binary function, where the opacity $1-T_k$ is the OR of the samples up to $k$. The product is this OR for independent samples, as in a participating medium, and the maximum for samples thresholded at one shared level, as along a ray through a solid object.

\begin{figure}[t]
\centering
\includegraphics[width=\columnwidth]{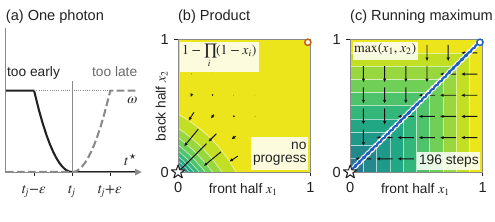}
\caption{\looseness=-1 \textbf{The too-early cost and its two extensions.} (a)~The cost of a photon at $t_j$ against the first crossing $\tst$ splits into too-early and too-late costs, capped at $\omega$ beyond $t_j\pm\varepsilon$. (b,\,c)~The too-early cost (darker is lower) of an eight-sample floater over the occupancies $x_1,x_2$ of its halves, both empty at the star. Gradient descent from a nearly opaque floater (dot) stalls for 20\,000 steps under the product and empties it in 196 under the running maximum.}
\label{fig:theory}
\end{figure}

\looseness=-1 With either transmittance, the too-early cost is a non-negative combination of the opacities, but the two differ in the gradient on a floater (\cref{fig:theory}b,\,c). Under the product, the derivative of an opacity with respect to the occupancy of one sample is the transmittance of the other samples, which nearly vanishes for a dense floater several samples thick, however many photons it blocks. Under the running maximum, the cost is convex with a non-decaying subgradient, and the floater is removed at a rate proportional to the photons it blocks. This argument does not cover per-bin losses, which act on a floater through the histogram it renders.

\subsection{Convex Bound on the Too-Late Cost}
\label{sec:late}

\looseness=-1 The too-late cost is a non-negative combination of the concave functions $1-\xb_k$ by summation by parts. A photon only indicates that some sample near its arrival lies inside the object. Any such sample $i$ is a witness, since $1-\xb_k\le\pos{1-x_{i}}$ for $i\le k$, where $\pos{\cdot}$ is the positive part. Each choice of witnesses thus bounds the concave term by a convex one. In place of the tangent of the concave--convex procedure~\cite{yuille2003cccp,narasimhan2005submodular}, we use a convex bound in the field, which does not guarantee monotone descent.

\looseness=-1 In the first iteration, the field does not yet show where the object is, and the thickness assumption provides the witness. A photon can be explained within the band $\varepsilon=\sqrt{2\omega}\,\sigma$ around it, at whose edge its cost reaches $\omega$ (\cref{fig:theory}a). If the object is thicker than this band along the ray, the point one band behind the detected return $\that_r$ (\cref{sec:read}) is inside it. In later iterations, we use the current field to place the witness between half a band and one band behind the return, at the distance $s_r$ that minimises \cref{eq:B}. For a ray with $N_r$ photons in its return and the witness $\bm{p}_r^{+}=\bm{r}(\that_r+s_r)$, the unexplained-photon cost $\omega N_r$ becomes
\begin{equation}
B_r(u)=\frac{\omega N_r}{m_r^{2}}\Big(\frac{\varepsilon}{s_r}\Big)^{2}\pos{\,\frac{m_r s_r}{\varepsilon}-u(\bm{p}_r^{+})}^{2}.
\label{eq:B}
\end{equation}
\looseness=-1 Here $m_r>0$ is the scale of $u$ that we define in \cref{sec:energy}. The term is at least $\omega N_r$ while the witness lies outside the object and vanishes once $u$ reaches $m_rs_r/\varepsilon$ there. It is thus a convex upper bound on that cost.

\subsection{Surface Localisation}
\label{sec:loc}

\looseness=-1 To localise the surface within the band, we take the occupancy near a photon at $t_j$ to be the ramp ${x(t)=\pos{(t-t_0+w)/(2w)}}$ of half-width $w$ about $t_0$, with offset $\delta=t_0-t_j$ and $|\delta|<w$, and ignore the cap $\omega$. The continuous too-early and too-late costs of this photon are then $(w-\delta)^3/(12w\sigma^2)$ and $(w+\delta)^3/(12w\sigma^2)$, whose sum is
\begin{equation}
\frac{\delta^2}{2\sigma^2}+\frac{w^2}{6\sigma^2}.
\label{eq:laplace}
\end{equation}
For every ramp width, the sum is the cost of a sharp crossing at $t_0$ plus a term that does not depend on the offset, which has two consequences. First, the slope $\partial_t u$ of $u$ along the ray sets the ramp width but not the location accuracy, and no eikonal term is required. Second, for a crossing at $\tst$, the $N_r$ photons of a return have, up to a constant, the cost of $N_r$ photons at their centroid $\that_r$. Linearising $u$ along the ray at $\bm{p}_r=\bm{r}(\that_r)$ gives the Gauss--Newton form that we discretise,
\begin{equation}
C_r(u)=\frac{N_r\,u(\bm{p}_r)^{2}}{2(\partial_t u)^{2}\sigma^{2}}.
\label{eq:C}
\end{equation}

\begin{table*}[t]
\centering\footnotesize\setlength{\tabcolsep}{5pt}
\caption{\looseness=-1 \textbf{Depth error (mm, $\downarrow$) on the test scenes of the Transientangelo (TA) protocol.} Ours is lower than TA in every pair. TA uses its released checkpoints, none below 10 PPP, and 1 and 0.1 PPP are thinned from 10 PPP. Rows use three views unless stated, and the object depth MAE excludes the back wall. $^\dagger$The boots checkpoint misses most of the object here in at least half of the novel views, and ours is also lower without boots. The TransientNeRF row averages the per-scene errors rounded to 1\,mm in~\cite{luo2025transientangelo}, with -- where none is reported.}
\label{tab:main}
\begin{tabular}{@{}l cc cc cc cc cc c c@{}}
\toprule
Flux (PPP) & \multicolumn{2}{c}{Full} & \multicolumn{2}{c}{300} & \multicolumn{2}{c}{150} & \multicolumn{2}{c}{50} & \multicolumn{2}{c}{10} & 1 & 0.1\\
\cmidrule(lr){2-3}\cmidrule(lr){4-5}\cmidrule(lr){6-7}\cmidrule(lr){8-9}\cmidrule(lr){10-11}\cmidrule(lr){12-12}\cmidrule(lr){13-13}
Scene & TA~\cite{luo2025transientangelo} & Ours & TA & Ours & TA & Ours & TA & Ours & TA & Ours & Ours & Ours\\
\midrule
baskets & 7.58 & \textbf{5.73} & 7.26 & \textbf{5.63} & 7.23 & \textbf{5.68} & 7.55 & \textbf{5.60} & 14.77 & \textbf{5.66} & 5.67 & 5.64\\
boots & 0.92 & \textbf{0.42} & 0.86 & \textbf{0.41} & 0.83 & \textbf{0.41} & 33.21$^\dagger$ & \textbf{0.41} & 0.62 & \textbf{0.42} & 0.43 & 0.58\\
chef & 3.88 & \textbf{0.91} & 3.50 & \textbf{0.95} & 4.81 & \textbf{0.92} & 2.73 & \textbf{0.94} & 5.17 & \textbf{0.94} & 1.08 & 1.76\\
food & 6.43 & \textbf{4.84} & 6.30 & \textbf{4.84} & 6.26 & \textbf{4.85} & 5.94 & \textbf{4.86} & 7.62 & \textbf{4.86} & 4.94 & 5.02\\
Mean & 4.70 & \textbf{2.97} & 4.48 & \textbf{2.96} & 4.78 & \textbf{2.97} & 12.36$^\dagger$ & \textbf{2.95} & 7.04 & \textbf{2.97} & 3.03 & 3.25\\
TransientNeRF~\cite{malik2023transientnerf} from~\cite{luo2025transientangelo} & \multicolumn{2}{c}{5.75} & \multicolumn{2}{c}{6.50} & \multicolumn{2}{c}{17.50} & \multicolumn{2}{c}{--} & \multicolumn{2}{c}{48.00} & -- & --\\
\midrule
Mean, 2 views & 6.34 & \textbf{3.07} & 3.92 & \textbf{3.11} & 5.09 & \textbf{3.06} & 16.47$^\dagger$ & \textbf{3.09} & 7.67 & \textbf{3.10} & 3.28 & 4.02\\
Mean, 5 views & 6.02 & \textbf{2.79} & 13.84$^\dagger$ & \textbf{2.80} & 5.36 & \textbf{2.80} & 7.99 & \textbf{2.81} & 6.54 & \textbf{2.81} & 2.82 & 2.95\\
Mean object depth MAE & 2.60 & \textbf{0.90} & 2.36 & \textbf{0.88} & 2.67 & \textbf{0.89} & 10.23$^\dagger$ & \textbf{0.88} & 4.94 & \textbf{0.89} & 0.95 & 1.17\\
\bottomrule
\end{tabular}
\end{table*}
\begin{figure*}[t]
\centering
{\setlength{\unitlength}{\dimexpr\textwidth*100/49500\relax}%
\begin{picture}(495,90.97)
\put(0,0){\includegraphics[width=\textwidth]{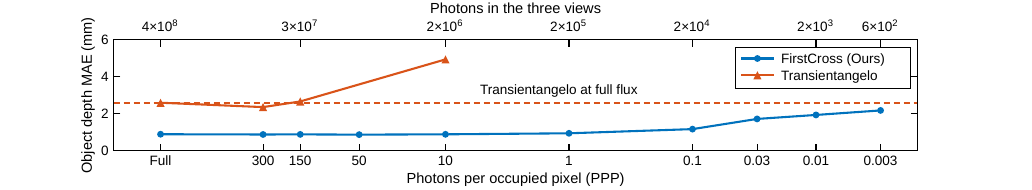}}%
\put(423.160,52.459){\makebox[0pt][l]{\fontencoding{T1}\fontfamily{phv}\fontseries{m}\fontshape{n}\fontsize{6.5}{8.4}\selectfont \cite{luo2025transientangelo}}}%
\end{picture}}
\caption{\looseness=-1 \textbf{Mean object depth MAE against photon count over the three-view test scenes.} Ours stays below Transientangelo's full-flux error (dashed) down to fewer than $10^3$ photons in total. Transientangelo at 50 PPP is omitted, as its boots checkpoint fails.}
\label{fig:curve}
\end{figure*}
\subsection{Overall Functional}
\label{sec:energy}

With a per-ray weight $z_r$ defined below, the functional is
\begin{equation}
\begin{aligned}
E(u)&=\sum\nolimits_r z_r\,\bar\ell_r(u)+\tfrac{\tau}{2}\langle u,L^{2}u\rangle,\\
\bar\ell_r(u)&=\sum\nolimits_k\Delta_{r,k}\,\xb_{r,k}+B_r+C_r,\\
x_{r,k}&=g\bigl((u(\bm{r}(\xi_k))+m_r)/m_r\bigr),\\
\Delta_{r,k}&=\ldn_r(\xi_k)-\ldn_r(\xi_{k+1}),\\
\ldn_r(t)&=\omega\,\pos{n_r(t)-q\,n_r}+a_r(t),
\end{aligned}
\label{eq:E}
\end{equation}
\looseness=-1 where $g(y)=2(\sqrt{1+\pos{y}^2}-1)$ is convex and non-decreasing, as the closure requires. The \emph{too-early term} carries the capped part $\ldn_r$ of the too-early cost. Here $n_r(t)$ counts the photons after $t+\varepsilon$ among all $n_r\ge N_r$ photons of the ray, $q$ is the fraction of photons allowed behind the crossing for multipath, mixed pixels and the IRF tail, and $a_r(t)$ is zero on rays with a return and otherwise the expected signal count of a crossing at $t$, which falls off with the square of $t$ (Sec.~S2). The \emph{too-late term} $B_r$ carries the capped part of the too-late cost, and the \emph{location term} $C_r$ carries the quadratic parts of both costs in the band around each return photon, summed in closed form by \cref{eq:laplace}. Outside the band the location term stays quadratic, as a cap would break convexity. The bending prior, with strength $\tau$ and discrete Laplacian $L$, fills in unobserved regions. All terms are convex in $u$ for fixed slopes, scales and witnesses.

\par\noindent\looseness=-1 \textbf{Scale of the field.} The likelihood depends only on the zero level set of $u$. The too-early and too-late terms fix the slope of $u$ along the ray at $m_r/\varepsilon$ (Sec.~S1). We set $m_r=\beta\varepsilon c_r$ with the incidence cosine $c_r$ of the ray, so that a plane of slope $\beta$ along its normal has slope $m_r/\varepsilon$ along every ray.

\par\noindent\looseness=-1 \textbf{Effective count.} Our likelihood treats the photons of a ray as independent draws from an exact model. Errors of calibration, pulse shape, residual pile-up and mixed pixels do not shrink with the photon count and can outweigh photon noise at full flux (Sec.~S5). We therefore coarsen the likelihood of each ray~\cite{miller2019coarsening}, which approximately raises it to the power $z_r=\alpha/(\alpha+n_r)$ (Sec.~S1). The effective count $z_rn_r$ stays below $\alpha$, and the functional stays convex.

\par\noindent\looseness=-1 \textbf{Parameters.} Measuring lengths in units of $\sigma$ makes the functional dimensionless, and the band $\varepsilon$ is fixed by $\omega$. From objects with $\sigma=4.4$\,mm to rooms with $\sigma=9$\,cm, we use $\omega=3.7$\,nats, $\beta=0.5$, $\tau=2.646$, $\alpha=20$ and $q=0.1$. The grid has $256^3$ voxels for objects and $384^3$ for rooms.

\subsection{Return Detection}
\label{sec:read}

\looseness=-1 The statistics in the functional are computed from the histograms with standard detection steps (Sec.~S2). On objects, each ray pools the photons of $4\times4$ pixels (one pixel on rooms), while depth is still rendered at every pixel. We take the return $\that_r$ of a ray as the centroid of its earliest strong matched-filter response, and a ray has a return if its local false-discovery rate against simulated background-only rays is at most $1/2$, with no threshold tuned per scene or flux. Rays without a return are treated as zero-photon rays. The full-flux captures are provided with object masks applied, and at low flux we use the masks of Transientangelo, giving both methods identical inputs.

\subsection{Solver and Rendering}
\label{sec:solver}

\looseness=-1 To minimise $E$, we alternate a dual update along rays with a linear solve for $u$. The dual update writes each running maximum as a maximum over distributions of where the ray enters the object and gives each sample a weight $W$. With these weights fixed, we replace each term by a quadratic model (Alg.~S1). We run 12 iterations from the empty grid $u=-\beta\varepsilon$ with unit slope and cosine. Each iteration is one damped dual update and one solve by FFT-preconditioned conjugate gradients, after which the slopes, cosines, scales and witnesses are updated. Keeping all four at their first-iteration values gives the single convex problem of \cref{tab:abl}.

\begin{figure*}[t]
\centering
{\setlength{\unitlength}{\dimexpr\textwidth*100/49500\relax}%
\begin{picture}(495,268.8)
\put(0,0){\includegraphics[width=495\unitlength]{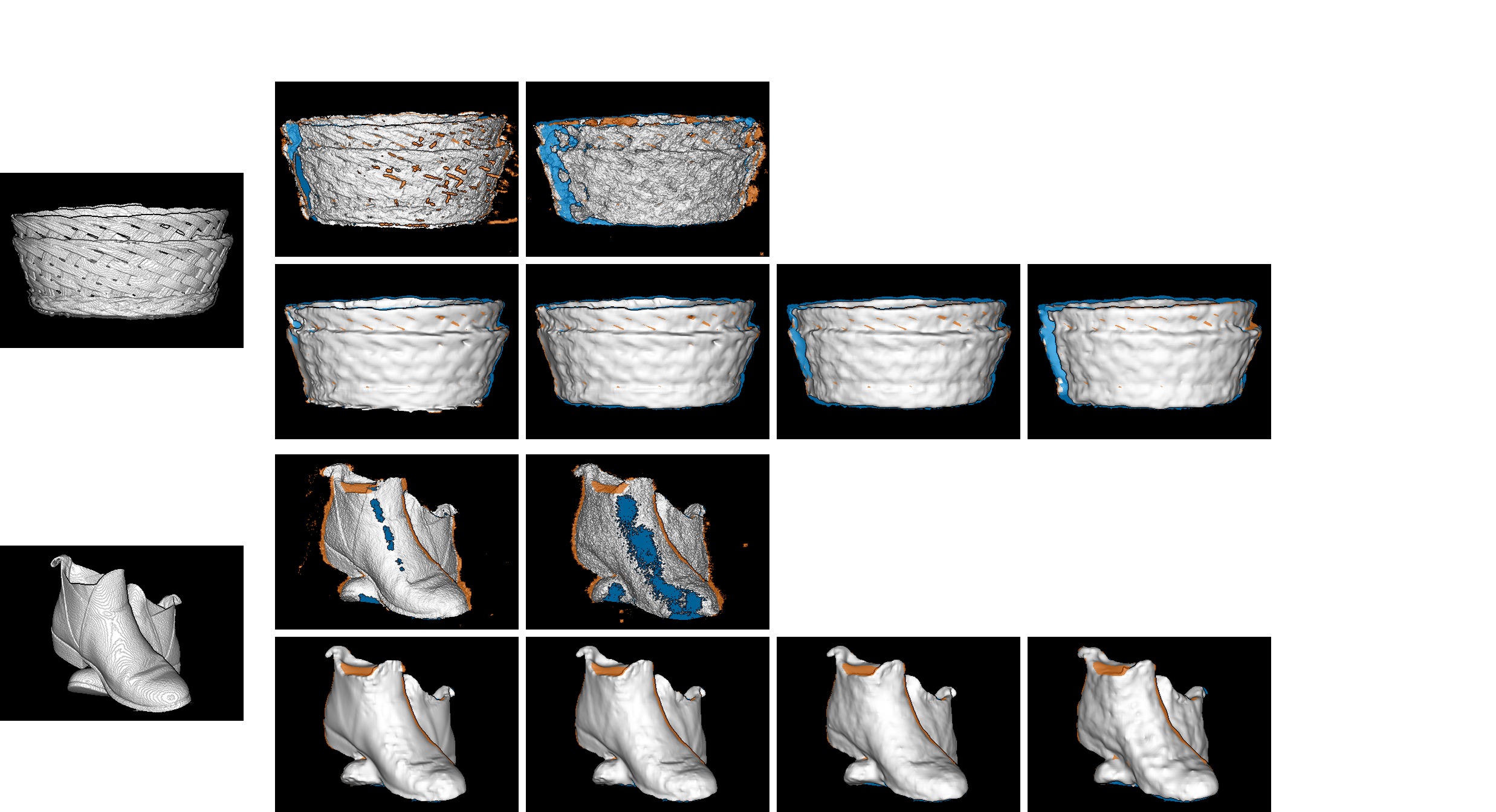}}%
\put(0,0){\includegraphics[width=495\unitlength]{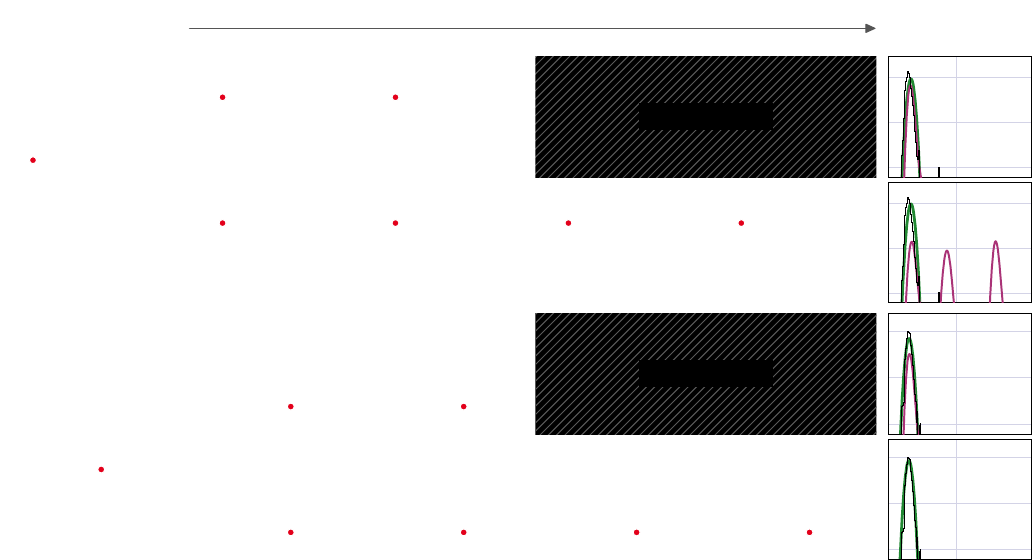}}%
\put(0,264.8){\makebox(0,0)[l]{\fontencoding{T1}\fontfamily{phv}\fontseries{m}\fontshape{n}\fontsize{7}{8.4}\selectfont \vphantom{Tg}\textcolor[RGB]{213,94,0}{\rule{4.4pt}{4.4pt}}~floaters\hspace{6pt}\textcolor[RGB]{0,114,178}{\rule{4.4pt}{4.4pt}}~holes\hspace{14pt}transients at the red dot (\textcolor[HTML]{E3001B}{\raisebox{.4pt}{\scalebox{.8}{$\bullet$}}}):\hspace{5pt}\textcolor[HTML]{000000}{\rule[1.75pt]{8pt}{.5pt}}~ground truth\hspace{7pt}\textcolor[HTML]{AA3377}{\rule[1.75pt]{10pt}{.9pt}}~Transientangelo~\cite{luo2025transientangelo}\hspace{7pt}\textcolor[HTML]{228833}{\rule[1.6pt]{10pt}{1.2pt}}~FirstCross (Ours)}}%
\put(255.8,255.2){\makebox(0,0)[c]{\fontencoding{T1}\fontfamily{phv}\fontseries{m}\fontshape{n}\fontsize{7}{8.4}\selectfont \setlength{\fboxsep}{1.2pt}\colorbox{white}{\strut fewer photons per pixel}}}%
\put(40.3,246.8){\makebox(0,0)[c]{\fontencoding{T1}\fontfamily{phv}\fontseries{m}\fontshape{n}\fontsize{7}{8.4}\selectfont \vphantom{Tg}ground truth}}%
\put(131.3,246.8){\makebox(0,0)[c]{\fontencoding{T1}\fontfamily{phv}\fontseries{m}\fontshape{n}\fontsize{7}{8.4}\selectfont \vphantom{Tg}full flux}}%
\put(214.3,246.8){\makebox(0,0)[c]{\fontencoding{T1}\fontfamily{phv}\fontseries{m}\fontshape{n}\fontsize{7}{8.4}\selectfont \vphantom{Tg}10 PPP}}%
\put(297.3,246.8){\makebox(0,0)[c]{\fontencoding{T1}\fontfamily{phv}\fontseries{m}\fontshape{n}\fontsize{7}{8.4}\selectfont \vphantom{Tg}1 PPP}}%
\put(380.3,246.8){\makebox(0,0)[c]{\fontencoding{T1}\fontfamily{phv}\fontseries{m}\fontshape{n}\fontsize{7}{8.4}\selectfont \vphantom{Tg}0.1 PPP}}%
\put(460.8,246.8){\makebox(0,0)[c]{\fontencoding{T1}\fontfamily{phv}\fontseries{m}\fontshape{n}\fontsize{7}{8.4}\selectfont \vphantom{Tg}transients}}%
\put(82.8,212.8){\rotatebox{90}{\makebox[0pt][c]{\fontencoding{T1}\fontfamily{phv}\fontseries{m}\fontshape{n}\fontsize{7}{8.4}\selectfont Transientangelo}}}%
\put(82.8,152.4){\rotatebox{90}{\makebox[0pt][c]{\fontencoding{T1}\fontfamily{phv}\fontseries{m}\fontshape{n}\fontsize{7}{8.4}\selectfont Ours}}}%
\put(40.3,216.8){\makebox(0,0)[c]{\fontencoding{T1}\fontfamily{phv}\fontseries{m}\fontshape{n}\fontsize{7}{8.4}\selectfont \vphantom{Tg}baskets, 2 views}}%
\put(82.8,89.4){\rotatebox{90}{\makebox[0pt][c]{\fontencoding{T1}\fontfamily{phv}\fontseries{m}\fontshape{n}\fontsize{7}{8.4}\selectfont Transientangelo}}}%
\put(82.8,29){\rotatebox{90}{\makebox[0pt][c]{\fontencoding{T1}\fontfamily{phv}\fontseries{m}\fontshape{n}\fontsize{7}{8.4}\selectfont Ours}}}%
\put(40.3,93.4){\makebox(0,0)[c]{\fontencoding{T1}\fontfamily{phv}\fontseries{m}\fontshape{n}\fontsize{7}{8.4}\selectfont \vphantom{Tg}boots, 3 views}}%
\put(494,212.47){\makebox(0,0)[r]{\fontencoding{T1}\fontfamily{phv}\fontseries{m}\fontshape{n}\fontsize{5.5}{6.6}\selectfont \vphantom{Tg}\textcolor[HTML]{555555}{10}}}%
\put(458.26,237.8){\makebox(0,0)[r]{\fontencoding{T1}\fontfamily{phv}\fontseries{m}\fontshape{n}\fontsize{5.5}{6.6}\selectfont \vphantom{Tg}\textcolor[HTML]{555555}{0.5\,ns}}}%
\put(493.8,237.6){\makebox(0,0)[r]{\fontencoding{T1}\fontfamily{phv}\fontseries{m}\fontshape{n}\fontsize{6}{7.2}\selectfont \vphantom{Tg}full flux}}%
\put(493.8,177.2){\makebox(0,0)[r]{\fontencoding{T1}\fontfamily{phv}\fontseries{m}\fontshape{n}\fontsize{6}{7.2}\selectfont \vphantom{Tg}10 PPP}}%
\put(493.8,114.2){\makebox(0,0)[r]{\fontencoding{T1}\fontfamily{phv}\fontseries{m}\fontshape{n}\fontsize{6}{7.2}\selectfont \vphantom{Tg}full flux}}%
\put(493.8,53.8){\makebox(0,0)[r]{\fontencoding{T1}\fontfamily{phv}\fontseries{m}\fontshape{n}\fontsize{6}{7.2}\selectfont \vphantom{Tg}10 PPP}}%
\put(493.8,47){\makebox(0,0)[r]{\fontencoding{T1}\fontfamily{phv}\fontseries{m}\fontshape{n}\fontsize{5.5}{6.6}\selectfont \vphantom{Tg}\textcolor[HTML]{AA3377}{Transientangelo:}}}%
\put(493.8,41){\makebox(0,0)[r]{\fontencoding{T1}\fontfamily{phv}\fontseries{m}\fontshape{n}\fontsize{5.5}{6.6}\selectfont \vphantom{Tg}\textcolor[HTML]{AA3377}{no return}}}%
\put(338.8,212.8){\makebox(0,0)[c]{\fontencoding{T1}\fontfamily{phv}\fontseries{m}\fontshape{n}\fontsize{6.5}{7.8}\selectfont \vphantom{Tg}\textcolor{white}{below its protocol}}}%
\put(338.8,89.4){\makebox(0,0)[c]{\fontencoding{T1}\fontfamily{phv}\fontseries{m}\fontshape{n}\fontsize{6.5}{7.8}\selectfont \vphantom{Tg}\textcolor{white}{below its protocol}}}%
\end{picture}}
\caption{\looseness=-1 \textbf{Novel views from full flux to 0.1 PPP, and transients on two test scenes.} The red dot marks a pixel where Transientangelo has a floater or hole (depth error over 2\,cm) at 10 PPP and both methods are within 5\,mm at full flux. There, Transientangelo's transient at 10 PPP is mostly late on baskets and missing on boots, whereas ours stays at the ground-truth peak. Transients, on a log scale, are scaled to the ground truth (measured at full flux). Grey lines mark powers of ten (horizontal) and 0.5\,ns after the ground-truth peak (vertical).}\label{fig:grid}
\end{figure*}
\par\noindent\looseness=-1 \textbf{Depth and intensity rendering.}\label{sec:readout} Depth is the first zero crossing of the field along the ray, refined by linear interpolation, or the location of its maximum if there is no zero crossing. Novel views and transients are rendered from the first crossing with a per-point maximum-likelihood albedo~$\rho$ (Sec.~S2) and without a learned appearance model.

\begin{figure}[t]
\centering
\includegraphics[width=\columnwidth]{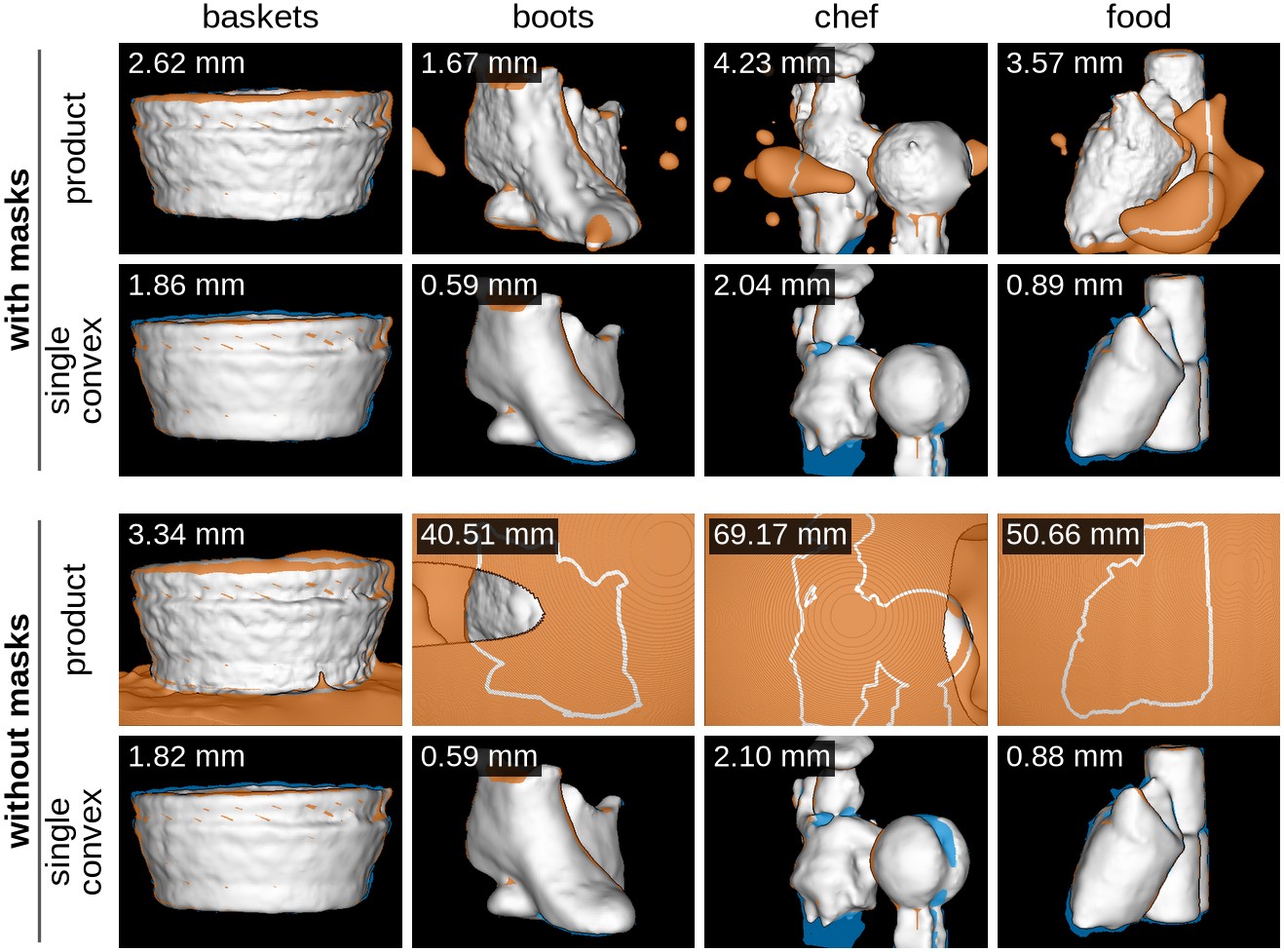}
\caption{\looseness=-1 \textbf{Product transmittance against the single convex problem.} At 0.1 PPP on the three-view test scenes, the product variant leaves floaters (orange) with masks, largest on chef and food. Without masks, it covers most of the view of boots, chef and food with spurious surfaces. Blue marks holes, and numbers are object depth MAEs.}
\label{fig:abl}
\end{figure}

\section{Experiments}
\label{sec:exp}

\par\noindent\looseness=-1 \textbf{Datasets and protocol.} We follow the protocol of Transientangelo~\cite{luo2025transientangelo}: its real captures, two, three or five input views, ten novel views, full flux and its low-flux settings of 300 to 10 PPP. Transientangelo results are obtained from its released checkpoints, which reproduce 14 of its 16 published three-view results (Sec.~S3). We further simulate 1 and 0.1 PPP, and down to 0.003 PPP in \cref{fig:curve}, by binomial thinning of 10 PPP, which keeps its signal-to-background ratio. The official depth error averages $|\hat d-d|$ over the frame, with zero outside the ground-truth mask. On baskets and food, this mask includes the turntable's back wall. We therefore also report the object depth MAE, which further zeroes the error beyond 0.30\,m from the scene origin. For room scale, we use the real rooms of SP-TransientBench~\cite{sptbench2026} at full flux and thinned to 1\%. All parameters are set on the validation scenes cinema, carving and library\_floor2. We report single runs, and Sec.~S5 adds repeated runs, other thinnings, extra ambient photons and camera rotations.

\par\noindent\looseness=-1 \textbf{Comparison to Transientangelo.} \Cref{tab:main} and Tab.~S4 show that ours has the lower official depth error on every test scene and flux level of the protocol with two, three and five views. With three views, our mean error is 2.97\,mm at both full flux and 10 PPP, whereas Transientangelo's rises from 4.70 to 7.04\,mm. This is a reduction of 37\% and 58\%, respectively, and Sec.~S2 gives bootstrap intervals. Our return detection accounts for much of the gap: TV-L1 fusion~\cite{zach2007tvl1} of our detected returns has less than half the object depth MAE of Transientangelo at full flux and 10 PPP (\cref{tab:main,tab:abl}). However, the best classical fusion has 1.3 to 1.8 times our mean object depth MAE at every flux level in Tab.~S6. On an RTX~4090, ours takes 51\,s in total on chef at 10 PPP, and Transientangelo trains for 1\,h\,26\,min.

\looseness=-1 In \cref{fig:grid} and Tab.~S8, the two methods differ most in holes and floaters, which dominate the mean error, but their medians are close at full flux. At 10 PPP, Transientangelo's median grows 2.7 and 5.4 times on chef and baskets, while ours changes by at most 7\%. Down to 1 PPP, three orders of magnitude below full flux, our object depth MAE grows by at most 19\% on any three-view scene (Tab.~S3). Our \mbox{Chamfer} distance is lower on every three-view test scene and protocol flux level, but Transientangelo has better completeness and F-score at 2\,mm in most of them (Sec.~S4).

\par\noindent\looseness=-1 \textbf{Lower photon counts.}\label{sec:below} In \cref{fig:curve} and Tab.~S6, our mean object depth MAE stays below Transientangelo's at full flux, and below TV-L1 fusion and screened Poisson reconstruction~\cite{kazhdan2013screened} of our detected returns, down to fewer than $10^3$ photons in three views. At the lowest flux, the reconstruction extends beyond the object, which the object depth MAE ignores (Sec.~S5). Without masks, our object depth MAE at 0.01 PPP is 2.04\,mm, against 1.94\,mm with masks and 9.39\,mm for the visual hull of the masks alone. Retrained with its official configuration, Transientangelo has a depth error of 5.48\,mm on chef at 10 PPP, close to the 5.17\,mm of its released checkpoint in \cref{tab:main}. At 1 PPP, however, it recovers no surface with or without adapted settings (Sec.~S3), whereas ours has 1.08\,mm. With its photons pooled as in ours, the adapted Transientangelo recovers chef and baskets at 1 PPP with 7.6 and 1.9 times our depth error.
\begin{figure}[t]
\centering
{\setlength{\unitlength}{\dimexpr\columnwidth*100/23625\relax}%
\begin{picture}(236.25,341.5)
\put(0,0){\includegraphics[width=236.25\unitlength]{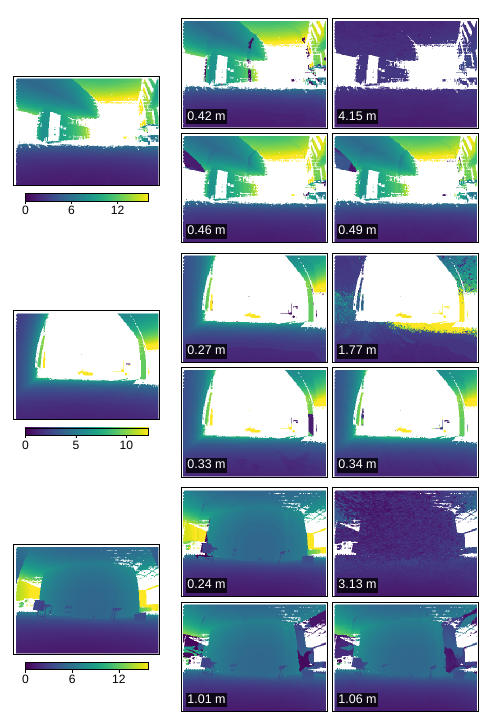}}%
\put(41.62,232.5){\makebox(0,0)[c]{\fontencoding{T1}\fontfamily{phv}\fontseries{m}\fontshape{n}\fontsize{6}{7.2}\selectfont \vphantom{Tg}depth (m)}}%
\put(41.62,310.2){\makebox(0,0)[c]{\fontencoding{T1}\fontfamily{phv}\fontseries{m}\fontshape{n}\fontsize{7}{8.4}\selectfont \vphantom{Tg}c4floor2}}%
\put(78.92,306.25){\rotatebox{90}{\makebox[0pt][c]{\fontencoding{T1}\fontfamily{phv}\fontseries{m}\fontshape{n}\fontsize{7}{8.4}\selectfont TransientNeRF}}}%
\put(78.92,251.25){\rotatebox{90}{\makebox[0pt][c]{\fontencoding{T1}\fontfamily{phv}\fontseries{m}\fontshape{n}\fontsize{7}{8.4}\selectfont Ours}}}%
\put(41.62,120){\makebox(0,0)[c]{\fontencoding{T1}\fontfamily{phv}\fontseries{m}\fontshape{n}\fontsize{6}{7.2}\selectfont \vphantom{Tg}depth (m)}}%
\put(41.62,197.7){\makebox(0,0)[c]{\fontencoding{T1}\fontfamily{phv}\fontseries{m}\fontshape{n}\fontsize{7}{8.4}\selectfont \vphantom{Tg}material\_building}}%
\put(78.92,193.75){\rotatebox{90}{\makebox[0pt][c]{\fontencoding{T1}\fontfamily{phv}\fontseries{m}\fontshape{n}\fontsize{7}{8.4}\selectfont TransientNeRF}}}%
\put(78.92,138.75){\rotatebox{90}{\makebox[0pt][c]{\fontencoding{T1}\fontfamily{phv}\fontseries{m}\fontshape{n}\fontsize{7}{8.4}\selectfont Ours}}}%
\put(41.62,7.5){\makebox(0,0)[c]{\fontencoding{T1}\fontfamily{phv}\fontseries{m}\fontshape{n}\fontsize{6}{7.2}\selectfont \vphantom{Tg}depth (m)}}%
\put(41.62,85.2){\makebox(0,0)[c]{\fontencoding{T1}\fontfamily{phv}\fontseries{m}\fontshape{n}\fontsize{7}{8.4}\selectfont \vphantom{Tg}physics\_building3}}%
\put(78.92,81.25){\rotatebox{90}{\makebox[0pt][c]{\fontencoding{T1}\fontfamily{phv}\fontseries{m}\fontshape{n}\fontsize{7}{8.4}\selectfont TransientNeRF}}}%
\put(78.92,26.25){\rotatebox{90}{\makebox[0pt][c]{\fontencoding{T1}\fontfamily{phv}\fontseries{m}\fontshape{n}\fontsize{7}{8.4}\selectfont Ours}}}%
\put(41.62,337.1){\makebox(0,0)[c]{\fontencoding{T1}\fontfamily{phv}\fontseries{m}\fontshape{n}\fontsize{7}{8.4}\selectfont \vphantom{Tg}ground truth}}%
\put(122.12,337.1){\makebox(0,0)[c]{\fontencoding{T1}\fontfamily{phv}\fontseries{m}\fontshape{n}\fontsize{7}{8.4}\selectfont \vphantom{Tg}full flux}}%
\put(194.62,337.1){\makebox(0,0)[c]{\fontencoding{T1}\fontfamily{phv}\fontseries{m}\fontshape{n}\fontsize{7}{8.4}\selectfont \vphantom{Tg}1\% flux}}%
{\color{red}\setlength{\fboxrule}{0.7pt}%
\put(162.52,299.95){\framebox(57.3,29.3){}}%
\put(162.52,179.58){\framebox(64.2,23.1){}}%
\put(162.52,74.43){\framebox(58.0,29.3){}}}%
\put(193.92,292.6){\makebox(0,0)[c]{\setlength{\fboxsep}{0.6pt}\colorbox{white}{\color{red}\fontencoding{T1}\fontfamily{phv}\fontseries{b}\fontshape{n}\fontsize{5}{6}\selectfont \vphantom{Tg}spurious surfaces}}}%
\put(203.02,208.98){\makebox(0,0)[c]{\setlength{\fboxsep}{0.6pt}\colorbox{white}{\color{red}\fontencoding{T1}\fontfamily{phv}\fontseries{b}\fontshape{n}\fontsize{5}{6}\selectfont \vphantom{Tg}noisy depth}}}%
\put(203.02,67.08){\makebox(0,0)[c]{\setlength{\fboxsep}{0.6pt}\colorbox{white}{\color{red}\fontencoding{T1}\fontfamily{phv}\fontseries{b}\fontshape{n}\fontsize{5}{6}\selectfont \vphantom{Tg}spurious surfaces}}}%
\end{picture}}
\caption{\looseness=-1 \textbf{Room-scale depth on the test rooms with three views.} At 1\% flux, TransientNeRF's depth becomes noisy with spurious surfaces in every room (red frames), whereas ours stays near its full-flux depth. Per room, we show the novel view where our 1\% flux error is the median, with one colour scale and white outside the benchmark mask. Numbers give the per-view mean depth MAE.}
\label{fig:rooms}
\end{figure}
\begin{table}[t]
\centering\footnotesize
\caption{\looseness=-1 \textbf{Ablation study.} Object depth MAE (mm, $\downarrow$) per flux level (PPP), averaged over the four three-view test scenes. Each variant changes one component and uses our solver. Product transmittance replaces the running maximum with NeRF's product. TV-L1 fusion uses our detected returns. Best in bold, second best underlined, and -- marks settings not run.}
\label{tab:abl}
\setlength{\tabcolsep}{4.2pt}
\begin{tabular}{@{}lcccc@{}}
\toprule
 & \multicolumn{3}{c}{With masks} & No masks\\
\cmidrule(lr){2-4}\cmidrule(l){5-5}
Method & Full & 10 & 0.1 & 10\\
\midrule
TV-L1 fusion~\cite{zach2007tvl1} & 1.18 & 1.61 & 2.03 & --\\
\midrule
Product transmittance & 1.12 & 1.11 & 3.02 & 12.46\\
Single convex problem & \underline{0.91} & \underline{0.92} & \underline{1.35} & \underline{0.91}\\
FirstCross (Ours) & \textbf{0.90} & \textbf{0.89} & \textbf{1.17} & \textbf{0.85}\\
\bottomrule
\end{tabular}

\end{table}

\par\noindent\looseness=-1 \textbf{Ablation study.} \Cref{tab:abl} compares our variants with TV-L1 fusion~\cite{zach2007tvl1}, the most accurate of three classical fusions down to 0.03 PPP (Tab.~S6). Compared with NeRF's product transmittance~\cite{mildenhall2020nerf}, the closure helps most at low flux, without masks and when the solver starts from an occupied grid. Even the single convex problem cuts the product variant's error at 0.1 PPP from 3.02 to 1.35\,mm and removes most of its floaters (\cref{fig:abl}). From an empty grid, this problem has a lower object depth MAE than Transientangelo on every three-view test scene and protocol flux level (Tab.~S15), and ours improves on it most at 0.1 PPP.

\looseness=-1 Without masks, the product variant produces spurious surfaces in front of the cameras (\cref{tab:abl,fig:abl}, Fig.~S13), with 2.5 to 7.3 times our error at 0.01 PPP (Tab.~S15). From the occupied grid $u=0$, it leaves free space filled even with masks, whereas down to 0.1 PPP this start changes the single convex problem's error by at most 0.17\,mm, due to inexact convergence. With L-BFGS from an empty grid, with or without masks, the product comes within 12\% of the closure's mean error. From the occupied grid $u=\beta\varepsilon$, however, the product under L-BFGS fills free space without masks at 0.01 PPP on three of four scenes, unlike the closure (Tab.~S16, Fig.~S15). Without the too-early term, the error averages 4.6 to 40 times ours down to 0.1 PPP on validation and test scenes, and without the too-late term, holes open where photons are few. When $q$, $\tau$ or $\alpha$ is halved or doubled, the mean object depth MAE on the validation objects changes by at most 2\%, and for $\omega$ or $\beta$, by at most 9\% (Tab.~S12).

\par\noindent\looseness=-1 \textbf{Room scale.} Across all rooms and view counts in \cref{tab:spt}, with unchanged dimensionless parameters, ours is 2.8 to 7.7 times more accurate than TransientNeRF at 1\% flux (\cref{fig:rooms}) and less accurate at full flux. We do not run Transientangelo on rooms, as its configuration assumes a bounding sphere of radius 0.4\,m and object masks at low flux. Ours takes 3\,min on library\_floor2, whereas TransientNeRF trains for 10 to 17\,h. On the test rooms, volumetric (TSDF) fusion~\cite{curless1996volumetric} of our detected returns has 1.2 to 4.5 times our mean error and outperforms TransientNeRF at 1\% flux (Sec.~S6).

\begin{table}[!t]
\centering\footnotesize
\caption{\looseness=-1 \textbf{Mean depth MAE (m, $\downarrow$) over all novel views of the SP-TransientBench test rooms.} Better of each pair in bold. TN (TransientNeRF) trains for the official 300k steps. On material\_building, our error drops at 1\% flux as the grid box around our returns shrinks. With the full-flux box, ours wins there (Sec.~S6).}
\label{tab:spt}
\setlength{\tabcolsep}{5pt}
\begin{tabular}{@{}lccccc@{}}
\toprule
& & \multicolumn{2}{c}{Full flux} & \multicolumn{2}{c}{1\% flux}\\
\cmidrule(lr){3-4}\cmidrule(l){5-6}
Room & Views & TN~\cite{malik2023transientnerf} & Ours & TN & Ours\\
\midrule
c4floor2 & 3 & \textbf{0.435} & 0.439 & 3.973 & \textbf{0.525}\\
 & 5 & \textbf{0.338} & 0.459 & 3.891 & \textbf{0.508}\\
material\_building & 3 & \textbf{0.382} & 0.705 & 2.135 & \textbf{0.325}\\
 & 5 & \textbf{0.319} & 0.718 & 1.767 & \textbf{0.315}\\
physics\_building3 & 3 & \textbf{0.768} & 1.209 & 3.303 & \textbf{1.193}\\
 & 5 & \textbf{0.697} & 1.066 & 3.455 & \textbf{1.122}\\
\bottomrule
\end{tabular}
\end{table}

\section{Conclusion}
\label{sec:limits}

\looseness=-1 We presented FirstCross, which reconstructs solid objects from a few views of single-photon lidar when photons are scarce. For a solid object, the Poisson likelihood of a ray with given signal and background rates is a function of its first crossing alone. Its too-early part has a tight convex closure, volume rendering with a running maximum, as does the non-increasing part of any first-crossing cost. Its too-late part has a convex upper bound under the thickness assumption. FirstCross therefore works through a short sequence of convex problems without a neural network, in about a minute per object. It has lower depth error than Transientangelo~\cite{luo2025transientangelo} in every setting of the original real-capture protocol. With three-view inputs thinned 100-fold below the protocol's lowest flux, its mean object depth MAE stays under half of Transientangelo's at full flux. FirstCross suits few-view 3D scanning of objects and rooms when acquisition must be fast. Code will be released.

\par\noindent\looseness=-1 \textbf{Limitations and Future Work.} The thickness assumption requires objects thicker than the band $\varepsilon$, about 24\,cm at room scale, and thin sheets break the first-iteration witness. Our field is also smooth at the scale of $\sigma$. At high flux, a neural surface is therefore sharper where views are dense, and on Transientangelo's three-view simulated scenes, our depth error is up to 1.36 times theirs above 10 PPP (Sec.~S7). There, because our solid fills space that no training view sees, our Chamfer distance is up to 32 times theirs. On two validation scenes, adding ten times as many ambient photons as were recorded raises our error at 0.1 PPP up to tenfold (Sec.~S5, which lists other limitations). Future work may refine the grid where views are dense, derive witnesses for thin sheets and estimate camera poses jointly. We also plan to extend the closure to other range sensors.

\clearpage
\section*{LLM Usage Statement}
Claude (Anthropic) was used under the authors' direction, through the Claude Code agent, to write and review the experiment code, run and monitor the experiments, and compute the statistics. It also maintained an append-only experiment log and drafted the text and figures of this paper, including this statement. Claude's code generated our simulated low-photon inputs, the added ambient photons and the camera rotations (\cref{sec:exp,sec:limits}, supplement S5 and~S7). The authors chose the research problem and the core method, approved each experiment plan before it was run, read every result, and edited the text. Gemini (Google) was used only to critique drafts. The authors take full responsibility for the content of this paper.

{\small
\bibliographystyle{ieeenat_fullname}
\bibliography{p2s}

\begin{thebibliography}{49}
\providecommand{\natexlab}[1]{#1}
\providecommand{\url}[1]{\texttt{#1}}
\expandafter\ifx\csname urlstyle\endcsname\relax
  \providecommand{\doi}[1]{doi: #1}\else
  \providecommand{\doi}{doi: \begingroup \urlstyle{rm}\Url}\fi

\bibitem[Bach(2013)]{bach2013learning}
Francis Bach.
\newblock Learning with submodular functions: A convex optimization
  perspective.
\newblock \emph{Foundations and Trends in Machine Learning}, 6\penalty0 (2-3),
  2013.

\bibitem[Calakli and Taubin(2011)]{calakli2011ssd}
Fatih Calakli and Gabriel Taubin.
\newblock {SSD}: Smooth signed distance surface reconstruction.
\newblock \emph{Computer Graphics Forum}, 30\penalty0 (7), 2011.

\bibitem[Carr et~al.(2001)Carr, Beatson, Cherrie, Mitchell, Fright, McCallum,
  and Evans]{carr2001rbf}
Jonathan~C. Carr, Richard~K. Beatson, Jon~B. Cherrie, Tim~J. Mitchell,
  W.~Richard Fright, Bruce~C. McCallum, and Tim~R. Evans.
\newblock Reconstruction and representation of {3D} objects with radial basis
  functions.
\newblock In \emph{Proc. SIGGRAPH}, 2001.

\bibitem[Curless and Levoy(1996)]{curless1996volumetric}
Brian Curless and Marc Levoy.
\newblock A volumetric method for building complex models from range images.
\newblock In \emph{Proc. SIGGRAPH}, 1996.

\bibitem[Dong et~al.(2026)Dong, Zhang, Wen, Qiang, Deng, Dong, Jiang, Li, Lu,
  Sun, Wang, Zheng, Shi, and Ren]{sptbench2026}
Hongzhou Dong, Zili Zhang, Ziting Wen, Yiheng Qiang, Runrong Deng, Wenle Dong,
  Ziwen Jiang, Xinyang Li, Rui Lu, Shuoyao Sun, Wenyu Wang, Haitao Zheng,
  Guodong Shi, and Xiaoqiang Ren.
\newblock {SP-TransientBench}: A real-captured single photon perception
  benchmark.
\newblock In \emph{Proc. European Conference on Computer Vision (ECCV)}, pages
  101--118, 2026.

\bibitem[Heide et~al.(2018)Heide, Diamond, Lindell, and
  Wetzstein]{heide2018sub}
Felix Heide, Steven Diamond, David~B. Lindell, and Gordon Wetzstein.
\newblock Sub-picosecond photon-efficient {3D} imaging using single-photon
  sensors.
\newblock \emph{Scientific Reports}, 8:\penalty0 17726, 2018.

\bibitem[Hou et~al.(2022)Hou, Wang, Wang, Qin, Qian, and He]{hou2022ipsr}
Fei Hou, Chiyu Wang, Wencheng Wang, Hong Qin, Chen Qian, and Ying He.
\newblock Iterative {P}oisson surface reconstruction ({iPSR}) for unoriented
  points.
\newblock \emph{ACM Transactions on Graphics}, 41\penalty0 (4), 2022.

\bibitem[Kazhdan and Hoppe(2013)]{kazhdan2013screened}
Michael Kazhdan and Hugues Hoppe.
\newblock Screened {P}oisson surface reconstruction.
\newblock \emph{ACM Transactions on Graphics}, 32\penalty0 (3), 2013.

\bibitem[Kazhdan et~al.(2006)Kazhdan, Bolitho, and Hoppe]{kazhdan2006poisson}
Michael Kazhdan, Matthew Bolitho, and Hugues Hoppe.
\newblock Poisson surface reconstruction.
\newblock In \emph{Proc. Eurographics Symposium on Geometry Processing}, 2006.

\bibitem[Kazhdan et~al.(2020)Kazhdan, Chuang, Rusinkiewicz, and
  Hoppe]{kazhdan2020envelope}
Michael Kazhdan, Ming Chuang, Szymon Rusinkiewicz, and Hugues Hoppe.
\newblock {P}oisson surface reconstruction with envelope constraints.
\newblock \emph{Computer Graphics Forum}, 39\penalty0 (5), 2020.

\bibitem[Kirmani et~al.(2014)Kirmani, Venkatraman, Shin, Cola{\c{c}}o, Wong,
  Shapiro, and Goyal]{kirmani2014first}
Ahmed Kirmani, Dheera Venkatraman, Dongeek Shin, Andrea Cola{\c{c}}o, Franco
  N.~C. Wong, Jeffrey~H. Shapiro, and Vivek~K. Goyal.
\newblock First-photon imaging.
\newblock \emph{Science}, 343\penalty0 (6166), 2014.

\bibitem[Kolev et~al.(2009)Kolev, Klodt, Brox, and
  Cremers]{kolev2009continuous}
Kalin Kolev, Maria Klodt, Thomas Brox, and Daniel Cremers.
\newblock Continuous global optimization in multiview {3D} reconstruction.
\newblock \emph{International Journal of Computer Vision}, 84\penalty0 (1),
  2009.

\bibitem[Kutulakos and Seitz(2000)]{kutulakos2000carving}
Kiriakos~N. Kutulakos and Steven~M. Seitz.
\newblock A theory of shape by space carving.
\newblock \emph{International Journal of Computer Vision}, 38\penalty0 (3),
  2000.

\bibitem[Labatut et~al.(2009)Labatut, Pons, and Keriven]{labatut2009robust}
Patrick Labatut, Jean-Philippe Pons, and Renaud Keriven.
\newblock Robust and efficient surface reconstruction from range data.
\newblock \emph{Computer Graphics Forum}, 28\penalty0 (8), 2009.

\bibitem[Lenain et~al.(2026)Lenain, Boyer, and Ponce]{lenain2026visibility}
Gabriel Lenain, Edmond Boyer, and Jean Ponce.
\newblock Surface reconstruction from range scans under visibility constraints.
\newblock \emph{arXiv preprint arXiv:2610.18005}, 2026.

\bibitem[Li et~al.(2025)Li, Sun, Hong, Xu, and Xiong]{li2025trt}
Yue Li, Shida Sun, Yu Hong, Feihu Xu, and Zhiwei Xiong.
\newblock {3D} reconstruction from transient measurements with time-resolved
  transformer.
\newblock \emph{arXiv preprint arXiv:2510.09205}, 2025.

\bibitem[Li et~al.(2023)Li, M{\"u}ller, Evans, Taylor, Unberath, Liu, and
  Lin]{li2023neuralangelo}
Zhaoshuo Li, Thomas M{\"u}ller, Alex Evans, Russell~H. Taylor, Mathias
  Unberath, Ming-Yu Liu, and Chen-Hsuan Lin.
\newblock Neuralangelo: High-fidelity neural surface reconstruction.
\newblock In \emph{Proc. IEEE/CVF Conference on Computer Vision and Pattern
  Recognition (CVPR)}, 2023.

\bibitem[Lin et~al.(2023)Lin, Xiao, Shi, and Wang]{lin2022pgr}
Siyou Lin, Dong Xiao, Zuoqiang Shi, and Bin Wang.
\newblock Surface reconstruction from point clouds without normals by
  parametrizing the {G}auss formula.
\newblock \emph{ACM Transactions on Graphics}, 42\penalty0 (2), 2023.

\bibitem[Lindell et~al.(2018)Lindell, O'Toole, and
  Wetzstein]{lindell2018single}
David~B. Lindell, Matthew O'Toole, and Gordon Wetzstein.
\newblock Single-photon {3D} imaging with deep sensor fusion.
\newblock \emph{ACM Transactions on Graphics}, 37\penalty0 (4), 2018.

\bibitem[Liu and Cooper(2010)]{liu2010ray}
Shubao Liu and David~B. Cooper.
\newblock Ray {Markov} random fields for image-based {3D} modeling: Model and
  efficient inference.
\newblock In \emph{Proc. IEEE Conference on Computer Vision and Pattern
  Recognition (CVPR)}, pages 1530--1537, 2010.

\bibitem[Lov{\'a}sz(1983)]{lovasz1983submodular}
L{\'a}szl{\'o} Lov{\'a}sz.
\newblock Submodular functions and convexity.
\newblock In \emph{Mathematical Programming: The State of the Art}. Springer,
  1983.

\bibitem[Luo et~al.(2025)Luo, Malik, and Lindell]{luo2025transientangelo}
Weihan Luo, Anagh Malik, and David~B. Lindell.
\newblock Transientangelo: Few-viewpoint surface reconstruction using
  single-photon lidar.
\newblock In \emph{Proc. IEEE/CVF Winter Conference on Applications of Computer
  Vision (WACV)}, 2025.

\bibitem[Malik et~al.(2023)Malik, Mirdehghan, Nousias, Kutulakos, and
  Lindell]{malik2023transientnerf}
Anagh Malik, Parsa Mirdehghan, Sotiris Nousias, Kiriakos~N. Kutulakos, and
  David~B. Lindell.
\newblock Transient neural radiance fields for lidar view synthesis and {3D}
  reconstruction.
\newblock In \emph{Advances in Neural Information Processing Systems
  (NeurIPS)}, 2023.

\bibitem[Malik et~al.(2024)Malik, Juravsky, Po, Wetzstein, Kutulakos, and
  Lindell]{malik2024flying}
Anagh Malik, Noah Juravsky, Ryan Po, Gordon Wetzstein, Kiriakos~N. Kutulakos,
  and David~B. Lindell.
\newblock Flying with photons: Rendering novel views of propagating light.
\newblock In \emph{Proc. European Conference on Computer Vision (ECCV)}, 2024.

\bibitem[Malik et~al.(2025)Malik, Attal, Xie, O'Toole, and
  Lindell]{malik2025inverse}
Anagh Malik, Benjamin Attal, Andrew Xie, Matthew O'Toole, and David~B. Lindell.
\newblock Neural inverse rendering from propagating light.
\newblock In \emph{Proc. IEEE/CVF Conference on Computer Vision and Pattern
  Recognition (CVPR)}, 2025.

\bibitem[Mildenhall et~al.(2020)Mildenhall, Srinivasan, Tancik, Barron,
  Ramamoorthi, and Ng]{mildenhall2020nerf}
Ben Mildenhall, Pratul~P. Srinivasan, Matthew Tancik, Jonathan~T. Barron, Ravi
  Ramamoorthi, and Ren Ng.
\newblock {NeRF}: Representing scenes as neural radiance fields for view
  synthesis.
\newblock In \emph{Proc. European Conference on Computer Vision (ECCV)}, 2020.

\bibitem[Miller et~al.(2024)Miller, Chen, Lai, and
  Gkioulekas]{miller2024objects}
Bailey Miller, Hanyu Chen, Alice Lai, and Ioannis Gkioulekas.
\newblock Objects as volumes: A stochastic geometry view of opaque solids.
\newblock In \emph{Proc. IEEE/CVF Conference on Computer Vision and Pattern
  Recognition (CVPR)}, 2024.

\bibitem[Miller and Dunson(2019)]{miller2019coarsening}
Jeffrey~W. Miller and David~B. Dunson.
\newblock Robust {B}ayesian inference via coarsening.
\newblock \emph{Journal of the American Statistical Association}, 114\penalty0
  (527):\penalty0 1113--1125, 2019.

\bibitem[Mu et~al.(2024)Mu, Sifferman, Jungerman, Li, Han, Gleicher, Gupta, and
  Li]{mu2024lowcost}
Fangzhou Mu, Carter Sifferman, Sacha Jungerman, Yiquan Li, Mark Han, Michael
  Gleicher, Mohit Gupta, and Yin Li.
\newblock Towards {3D} vision with low-cost single-photon cameras.
\newblock In \emph{Proc. IEEE/CVF Conference on Computer Vision and Pattern
  Recognition (CVPR)}, 2024.

\bibitem[Narasimhan and Bilmes(2005)]{narasimhan2005submodular}
Mukund Narasimhan and Jeff Bilmes.
\newblock A submodular-supermodular procedure with applications to
  discriminative structure learning.
\newblock In \emph{Proc. Conference on Uncertainty in Artificial Intelligence
  (UAI)}, 2005.

\bibitem[Oechsle et~al.(2021)Oechsle, Peng, and Geiger]{oechsle2021unisurf}
Michael Oechsle, Songyou Peng, and Andreas Geiger.
\newblock {UNISURF}: Unifying neural implicit surfaces and radiance fields for
  multi-view reconstruction.
\newblock In \emph{Proc. IEEE/CVF International Conference on Computer Vision
  (ICCV)}, 2021.

\bibitem[Peng et~al.(2020)Peng, Xiong, Huang, Li, Liu, and Xu]{peng2020photon}
Jiayong Peng, Zhiwei Xiong, Xin Huang, Zheng-Ping Li, Dong Liu, and Feihu Xu.
\newblock Photon-efficient {3D} imaging with a non-local neural network.
\newblock In \emph{Proc. European Conference on Computer Vision (ECCV)}, 2020.

\bibitem[Peng et~al.(2021)Peng, Jiang, Liao, Niemeyer, Pollefeys, and
  Geiger]{peng2021sap}
Songyou Peng, Chiyu Jiang, Yiyi Liao, Michael Niemeyer, Marc Pollefeys, and
  Andreas Geiger.
\newblock Shape as points: A differentiable {P}oisson solver.
\newblock In \emph{Advances in Neural Information Processing Systems
  (NeurIPS)}, 2021.

\bibitem[Rapp and Goyal(2017)]{rapp2017few}
Joshua Rapp and Vivek~K. Goyal.
\newblock A few photons among many: Unmixing signal and noise for
  photon-efficient active imaging.
\newblock \emph{IEEE Transactions on Computational Imaging}, 3\penalty0 (3),
  2017.

\bibitem[Savinov et~al.(2015)Savinov, Ladick{\'y}, H{\"a}ne, and
  Pollefeys]{savinov2015ray}
Nikolay Savinov, L'ubor Ladick{\'y}, Christian H{\"a}ne, and Marc Pollefeys.
\newblock Discrete optimization of ray potentials for semantic {3D}
  reconstruction.
\newblock In \emph{Proc. IEEE Conference on Computer Vision and Pattern
  Recognition (CVPR)}, 2015.

\bibitem[Savinov et~al.(2016)Savinov, H{\"a}ne, Ladick{\'y}, and
  Pollefeys]{savinov2016semantic}
Nikolay Savinov, Christian H{\"a}ne, L'ubor Ladick{\'y}, and Marc Pollefeys.
\newblock Semantic {3D} reconstruction with continuous regularization and ray
  potentials using a visibility consistency constraint.
\newblock In \emph{Proc. IEEE Conference on Computer Vision and Pattern
  Recognition (CVPR)}, 2016.

\bibitem[Scheuble et~al.(2025)Scheuble, Ramazzina, Holzh{\"u}ter, Gasperini,
  Peters, Tombari, Bijelic, and Heide]{scheuble2025lasso}
Dominik Scheuble, Andrea Ramazzina, Hanno Holzh{\"u}ter, Stefano Gasperini,
  Steven Peters, Federico Tombari, Mario Bijelic, and Felix Heide.
\newblock Transient {LASSO}: Transient large-scale scene reconstruction.
\newblock In \emph{Proc. SIGGRAPH Asia Conference Papers}, 2025.

\bibitem[Shin et~al.(2015)Shin, Kirmani, Goyal, and Shapiro]{shin2015photon}
Dongeek Shin, Ahmed Kirmani, Vivek~K. Goyal, and Jeffrey~H. Shapiro.
\newblock Photon-efficient computational {3-D} and reflectivity imaging with
  single-photon detectors.
\newblock \emph{IEEE Transactions on Computational Imaging}, 1\penalty0 (2),
  2015.

\bibitem[Speierer and Jarabo(2026)]{speierer2026nonexp}
S{\'e}bastien Speierer and Adrian Jarabo.
\newblock Generalized non-exponential {G}aussian splatting.
\newblock \emph{arXiv preprint arXiv:2603.02887}, 2026.

\bibitem[Tachella et~al.(2019)Tachella, Altmann, Mellado, McCarthy, Tobin,
  Buller, Tourneret, and McLaughlin]{tachella2019real}
Juli{\'a}n Tachella, Yoann Altmann, Nicolas Mellado, Aongus McCarthy, Rachael
  Tobin, Gerald~S. Buller, Jean-Yves Tourneret, and Stephen McLaughlin.
\newblock Real-time {3D} reconstruction from single-photon lidar data using
  plug-and-play point cloud denoisers.
\newblock \emph{Nature Communications}, 10:\penalty0 4984, 2019.

\bibitem[Tagliasacchi et~al.(2011)Tagliasacchi, Olson, Zhang, Hamarneh, and
  Cohen-Or]{tagliasacchi2011vase}
Andrea Tagliasacchi, Matt Olson, Hao Zhang, Ghassan Hamarneh, and Daniel
  Cohen-Or.
\newblock {VASE}: Volume-aware surface evolution for surface reconstruction
  from incomplete point clouds.
\newblock \emph{Computer Graphics Forum}, 30\penalty0 (5), 2011.

\bibitem[Ulusoy et~al.(2015)Ulusoy, Geiger, and Black]{ulusoy2015ray}
Ali~Osman Ulusoy, Andreas Geiger, and Michael~J. Black.
\newblock Towards probabilistic volumetric reconstruction using ray potentials.
\newblock In \emph{Proc. International Conference on {3D} Vision ({3DV})},
  2015.

\bibitem[Vicini et~al.(2021)Vicini, Jakob, and Kaplanyan]{vicini2021nonexp}
Delio Vicini, Wenzel Jakob, and Anton Kaplanyan.
\newblock A non-exponential transmittance model for volumetric scene
  representations.
\newblock \emph{ACM Transactions on Graphics}, 40\penalty0 (4), 2021.

\bibitem[Wang et~al.(2021)Wang, Liu, Liu, Theobalt, Komura, and
  Wang]{wang2021neus}
Peng Wang, Lingjie Liu, Yuan Liu, Christian Theobalt, Taku Komura, and Wenping
  Wang.
\newblock {NeuS}: Learning neural implicit surfaces by volume rendering for
  multi-view reconstruction.
\newblock In \emph{Advances in Neural Information Processing Systems
  (NeurIPS)}, 2021.

\bibitem[Wen et~al.(2026)Wen, Deng, Zhang, Zheng, Xu, Ren, Shi, and
  Ding]{wen2026mixed}
Ziting Wen, Runrong Deng, Zili Zhang, Haitao Zheng, Yuecong Xu, Xiaoqiang Ren,
  Guodong Shi, and Kemi Ding.
\newblock Resolving mixed single-photon {LiDAR} returns for foreground-view and
  hidden scene reconstruction.
\newblock \emph{arXiv preprint arXiv:2610.01206}, 2026.

\bibitem[Yariv et~al.(2020)Yariv, Kasten, Moran, Galun, Atzmon, Basri, and
  Lipman]{yariv2020idr}
Lior Yariv, Yoni Kasten, Dror Moran, Meirav Galun, Matan Atzmon, Ronen Basri,
  and Yaron Lipman.
\newblock Multiview neural surface reconstruction by disentangling geometry and
  appearance.
\newblock In \emph{Advances in Neural Information Processing Systems
  (NeurIPS)}, 2020.

\bibitem[Yariv et~al.(2021)Yariv, Gu, Kasten, and Lipman]{yariv2021volsdf}
Lior Yariv, Jiatao Gu, Yoni Kasten, and Yaron Lipman.
\newblock Volume rendering of neural implicit surfaces.
\newblock In \emph{Advances in Neural Information Processing Systems
  (NeurIPS)}, 2021.

\bibitem[Yuille and Rangarajan(2003)]{yuille2003cccp}
Alan~L. Yuille and Anand Rangarajan.
\newblock The concave-convex procedure.
\newblock \emph{Neural Computation}, 15\penalty0 (4), 2003.

\bibitem[Zach et~al.(2007)Zach, Pock, and Bischof]{zach2007tvl1}
Christopher Zach, Thomas Pock, and Horst Bischof.
\newblock A globally optimal algorithm for robust {TV-L1} range image
  integration.
\newblock In \emph{Proc. IEEE International Conference on Computer Vision
  (ICCV)}, 2007.

\end{thebibliography}
}

\end{document}